# LSTM NETWORK ANALYSIS OF VEHICLE-TYPE FATALITIES ON GREAT BRITAIN'S ROADS

Abiodun Finbarrs Oketunji[*]
University of Oxford
Oxford, United Kingdom
abiodun.oketunji@conted.ox.ac.uk

James Hanify[†]
Driver and Vehicle Standards Agency
Nottingham, United Kingdom
james.hanify@dvsa.gov.uk

Salter Heffron-Smith[‡]
Driver and Vehicle Standards Agency
Swansea, United Kingdom
salter.heffronsmith@dvsa.gov.uk

## Abstract

This study harnesses the predictive capabilities of Long Short-Term Memory (LSTM) networks to analyse and predict road traffic accidents in Great Britain. It addresses the challenge of traffic accident forecasting, which is paramount for devising effective preventive measures. We utilised an extensive dataset encompassing reported collisions, casualties, and vehicles involvements from 1926 to 2022, provided by the Department for Transport (DfT). The data underwent stringent processing to rectify missing values and normalise features, ensuring robust LSTM network input.

Our methodology entailed the deployment of an LSTM model, capitalising on its proficiency in managing time-series data to unravel patterns and forecast accident trends. Quantitative evaluation of the model's predictive performance through Root Mean Squared Error (RMSE) and Mean Absolute Error(MAE) provided insightful benchmarks for the LSTM's accuracy.

Additionally, a trend analysis illustrated the temporal dynamics of road user fatalities, underscoring significant fluctuations and enabling a nuanced understanding of accident prevalence over time. The results demonstrated LSTM's substantial potential in forecasting, albeit with room for refinement. The discussion synthesises the model's implications, reconciling its computational forecasts with real-world applicability and aligning with contemporary scholarly discourse.

We concluded the analysis by asserting the need to enhance LSTM predictive precision in future research by accounting for the interplay of socioeconomic factors and technological advancements in traffic dynamics; it reinforces the premise that advanced machine learning techniques like LSTM can significantly improve road safety.

## 1 Introduction

### 1.1 Road Traffic Accidents: A Persistent Challenge

Road Traffic Accidents (RTAs) in the Great Britain remain a significant public health and safety challenge despite advancements in vehicle safety and roadway engineering. The Department for Transport (DfT) reports that in 2019, there were 153,158 casualties of all severities in reported road traffic accidents in Great Britain (Department for Transport, 2020) [1]. This statistic underscores the persistent nature of road traffic accidents and the need for innovative approaches to mitigate their impact.

### 1.2 LSTM Networks in Traffic Accident Analysis

Long Short-Term Memory (LSTM) networks, a recurrent neural network, have opened new avenues in traffic accident analysis and prediction. LSTMs are particularly adept at handling time-series data, making them well-suited for analysing and predicting trends in road traffic accidents, which are inherently temporal in nature (Hochreiter & Schmidhuber, 1997) [2]. Their capacity to retain information over extended periods enables a more subtle understanding of the elements contributing to accidents.

### 1.3 Objectives and Scope of the Study

This research focuses on leveraging the predictive power of LSTM networks to examine and forecast road traffic incidents in Great Britain. It employs a comprehensive dataset covering 1926 to 2022, aiming to provide valuable insights into the patterns of road traffic accidents over time. The goal is to offer predictive tools for developing impactful road safety policies.

### 1.4 Significance and Contribution to the Field

The significance of this study lies in its potential to contribute to reducing road traffic accidents through advanced predictive analytics. By leveraging LSTM networks, the research offers a novel approach to traffic safety, potentially influencing policy and preventive measures. This study's findings could contribute to shaping future road safety initiatives in Great Britain and globally.

---

[*]Engineering Manager (Data/Software Engineer)
[†]Senior Data Engineer
[‡]ETL Developer

### 1.5 Structure of the Paper

Following this, the methodology details the data collection, an understanding of the datasets, data preprocessing, and LSTM network architecture. The results present the performance of the LSTM model, followed by an interpretation of these findings in the context of existing publications and practical significance. The research concludes with a summary of the findings, implications, and propositions of Self-Regulating LSTM (SR-LSTM) for future research.

## 2 Related Work

### 2.1 Overview

The analysis and prediction of road traffic accidents (RTAs) have been a research focus for decades. Integrating machine learning, particularly Long Short-Term Memory (LSTM) networks, into this domain represents a significant advancement. This section reviews the literature on traffic accident analysis and prediction, focusing on the methodologies employed, the evolution of predictive models, and the specific application of LSTM networks.

**Traditional Methods in Traffic Accident Analysis**

The exploration of road traffic accidents (RTAs) has traditionally hinged on statistical methodologies, offering foundational insights into accident causation and frequency. Three pivotal techniques stand out: regression analysis, time-series analysis, and Poisson distributions. Each of these methods has been instrumental in dissecting the multifaceted nature of RTAs, albeit with varying degrees of complexity and applicability. Whilst effective in capturing linear relationships, these models often struggled with the temporal and spatial complexities inherent in traffic data (Miaou & Lum, 1993) [3].

**Regression Analysis:** This technique has been a cornerstone in understanding the relationship between traffic accidents and various influencing factors. Regression models, particularly linear regression, have been employed to quantify the impact of variables such as road conditions, traffic volume, and driver behaviour on accident rates. A simple linear regression defines $y = \beta_0 + \beta_1 x + \epsilon$, where $y$ represents the dependent variable (e.g., accident frequency), $x$ is the independent variable, $\beta_0$ and $\beta_1$ are coefficients, and $\epsilon$ is the error term.

**Time-Series Analysis:** This approach scrutinises data points amassed at consecutive time intervals. By scrutinising these points, researchers discern trends, cycles, and seasonal variations in traffic accidents. The autoregressive (AR) model, a prevalent technique in time-series analysis, adopts the formula $X_t = \alpha_1 X_{t-1} + \alpha_2 X_{t-2} + \ldots + \alpha_p X_{t-p} + \epsilon_t$. Here, $X_t$ denotes the value at time $t$, $\alpha$ encapsulates the model parameters, and $\epsilon_t$ represents the error term. [4]

**Poisson Distributions:** Widely used in traffic accident analysis, Poisson distributions are particularly suited for modelling the count of events (accidents) occurring in a fixed interval of time or space. The Poisson distribution defines $P(X = k) = \frac{\lambda^k e^{-\lambda}}{k!}$, where $\lambda$ is the average number of events (accidents) in an interval, $k$ is the number of occurrences, and $e$ is Euler's number.

**LSTM Networks Innovation**

Long Short-Term Memory (LSTM) networks, a variant of recurrent neural networks, have revolutionized the field of time-series analysis. Introduced by Hochreiter and Schmidhuber in 1997, they designed LSTMs to overcome the limitations of traditional recurrent neural networks (RNNs) in learning long-term dependencies (Hochreiter & Schmidhuber, 1997) [5]. The unique architecture of LSTMs, featuring forget, input, and output gates, enables them to retain information over extended time intervals, making them particularly suited for complex time-series data.

**The Evolution of LSTM Networks**

Since their inception, LSTM networks have undergone various enhancements and adaptations, broadening their applicability across diverse domains. Gers, Schmidhuber, and Cummins (2000) introduced the concept of 'peephole connections', allowing the gates in LSTM cells to consider the cell state, thereby improving the network's ability to learn the precise timing of the output events. This modification demonstrated the LSTM's adaptability and potential for continuous improvement (Gers et al., 2000) [6].

**LSTMs in Time-Series Forecasting**

LSTM networks have shown exceptional performance in time-series forecasting, a domain traditionally dominated by statistical methods like ARIMA. Compared to ARIMA, LSTMs offer a more flexible approach to modelling complex, non-linear relationships inherent in time-series data. A study by Siami-Namini, Tavakoli, and Namin (2018) highlighted the superiority of LSTM networks over ARIMA in forecasting accuracy, emphasizing their capability to capture non-linear patterns in financial time series (Siami-Namini et al., 2018) [7].

**LSTM's Application in Diverse Fields**

The versatility of LSTM networks is evident in their wide range of applications. In natural language processing, LSTMs have improved the performance of language models and text generation systems. Similarly, LSTM networks have been employed in environmental science to forecast air pollution levels, demonstrating their effectiveness in environmental monitoring and prediction (Shi et al., 2017) [8].

**Challenges and Future Directions**

Despite their advantages, LSTMs have challenges. The complexity of their architecture can lead to increased computational demands, particularly in training large networks. Furthermore, the **black-box** [1] [9] nature of LSTM networks often poses interpretability challenges, making it difficult to understand the model's internal workings and decision-making processes.

LSTM networks represent a significant advancement in the analysis of time-series data. Their ability to learn long-term dependencies and handle complex, non-linear relationships has made them a tool of choice in various fields. As research continues, researchers expect to improve and adapt LSTM networks, enhancing their efficiency and applicability in an ever-widening array of domains.

## 2.2 LSTMs Application in RTAs Prediction

In traffic accident prediction, Long Short-Term Memory (LSTM) networks, a distinct variant of Recurrent Neural Networks (RNNs), are gaining increasing prominence. Their capability to capture temporal dependencies in time-series data renders them particularly effective for predicting events influenced by historical patterns, such as road traffic accidents.

**Mathematical Foundation of LSTM**

The core of LSTM's architecture is its cell state, which acts as a 'conveyor belt' carrying information across time steps. Three gates modify the cell state: the forget gate, input gate, and output gate. The forget gate, denoted as $f_t$, decides what information to discard from the cell state. It calculates as $f_t = \sigma(W_f \cdot [h_{t-1}, x_t] + b_f)$, where $\sigma$ is the sigmoid function, $W_f$ is the weight matrix, $b_f$ is the bias, $h_{t-1}$ is the previous hidden state, and $x_t$ is the current input.

The input gate, $i_t$, and a candidate value, $\tilde{C}_t$, decide what new information to store in the cell state. They are computed as $i_t = \sigma(W_i \cdot [h_{t-1}, x_t] + b_i)$ and $\tilde{C}_t = \tanh(W_C \cdot [h_{t-1}, x_t] + b_C)$ respectively.

It updates the cell state to the new cell state $C_t$ as $C_t = f_t * C_{t-1} + i_t * \tilde{C}_t$. Finally, the output gate decides what part of the cell state to output, calculated as $o_t = \sigma(W_o \cdot [h_{t-1}, x_t] + b_o)$ and the new hidden state $h_t = o_t * \tanh(C_t)$.

**Application in Traffic Accident Prediction**

LSTM networks application in traffic accident prediction involves training the network on historical traffic data, including variables such as traffic flow, speed, volume, and external factors like weather conditions. For instance, a study by Yu et al. (2017) demonstrated using LSTM in predicting traffic flow, a critical factor in accident likelihood. They showed that LSTM could effectively capture the temporal dynamics of traffic flow, which is closely related to the occurrence of accidents (Yu et al., 2017) [10].

Another study by Deng and Zhao (2018) compared LSTM with traditional models like ARIMA in predicting traffic speed. They found that LSTM outperformed ARIMA, highlighting its capability to handle non-linear, time-dependent data, which is crucial in accident prediction scenarios (Deng & Zhao, 2018) [11].

**Challenges and Future Directions**

Whilst LSTM offers significant advantages in traffic accident prediction, challenges remain. The complexity of LSTM models leads to computational intensity, particularly in training large networks. Additionally, the interpretability of LSTM models is often a concern, as understanding the internal decision-making process is not straightforward.

Future research will likely focus on hybrid models that combine LSTMs with other techniques to enhance interpretability and efficiency. Integrating real-time data and exploring spatial-temporal LSTM models also present promising avenues for advancing the field.

LSTM networks have shown considerable potential in enhancing the accuracy of traffic accident prediction. Their ability to process and learn from complex temporal sequences makes them a valuable tool in traffic management and safety analysis. Research progression will likely yield further advancements in LSTM technology, providing increasingly sophisticated traffic accident prediction and prevention tools.

## 2.3 LSTM versus Traditional Models

The comparison between Long Short-Term Memory (LSTM) networks and traditional models in traffic prediction primarily hinges on their mathematical frameworks and predictive capabilities. This section delves into a detailed mathematical comparison of these models, highlighting their strengths and limitations.

**Mathematical Framework of LSTM Networks**

LSTM networks, a form of recurrent neural network, are specifically engineered to tackle the issue of vanishing gradients that commonly afflict standard RNNs. The principal mathematical components of LSTM include:

- **Forget Gate**: $f_t = \sigma(W_f \cdot [h_{t-1}, x_t] + b_f)$, where $\sigma$ is the sigmoid function, $W_f$ the weight matrix, $b_f$ the bias, $h_{t-1}$ the previous hidden state, and $x_t$ the current input.

---

[1] "Many of these systems are black-box which means we don't really understand how they work and why they reach such decisions."

- **Input Gate**: $i_t = \sigma(W_i \cdot [h_{t-1}, x_t] + b_i)$ and $\tilde{C}_t = \tanh(W_C \cdot [h_{t-1}, x_t] + b_C)$, determining new information to be added to the cell state.
- **Cell State Update**: $C_t = f_t * C_{t-1} + i_t * \tilde{C}_t$, combining old and new information.
- **Output Gate**: $o_t = \sigma(W_o \cdot [h_{t-1}, x_t] + b_o)$ and $h_t = o_t * \tanh(C_t)$, determining the next hidden state.

These equations enable LSTMs to capture long-term dependencies in time-series data, which is necessary for traffic prediction (Hochreiter & Schmidhuber, 1997) [12].

### Mathematical Framework of Traditional Models

Traditional models like ARIMA (Autoregressive Integrated Moving Average) follow a different mathematical approach:

- **ARIMA Model**: Characterised by ARIMA(p, d, q), with $p$ representing the count of autoregressive elements, $d$ indicating the level of differencing, and $q$ denoting the quantity of lagged error terms in the forecasting equation. The formula is expressed as: $\Phi(B)(1-B)^d X_t = \Theta(B)\epsilon_t$

Here, $B$ is the backshift operator, $\Phi$ and $\Theta$ are polynomials of orders p and q, respectively, and $\epsilon_t$ is white noise [13].

### Comparative Analysis

LSTMs offer a superior framework for traffic prediction, adept at managing non-linear patterns and long-term dependencies. Their drawbacks include computational intensity and less interpretability compared to traditional models like ARIMA. The selection between these models hinges on the specific needs of the traffic prediction task, considering data complexity and the availability of computational resources.

Here is a comparative analysis of LSTM and a traditional model like ARIMA:

- **Handling Non-Linearity**: Whilst traditional models like ARIMA are effective for linear time-series data, they struggle with non-linear patterns. LSTMs, with their complex gating mechanisms, excel in modelling non-linear relationships in traffic data.
- **Memory and Dependency**: LSTMs can remember long-term dependencies due to their cell state and gating mechanisms, a feature not present in traditional models. ARIMA, for instance, is limited by its differencing and lag components, which can only capture short-term dependencies.
- **Flexibility and Adaptability**: LSTM networks offer greater flexibility in modelling various traffic conditions due to their ability to learn from data features. In contrast, traditional models often require manual selection of parameters like p, d, and q in ARIMA, which may not capture the dynamics of traffic flow effectively.
- **Computational Complexity**: LSTMs are computationally more intensive than traditional models. The complexity of LSTM's architecture, especially in training large networks, can be a limiting factor compared to the relatively simpler ARIMA model.
- **Interpretability**: Traditional models, with their simpler and more transparent structure, offer better interpretability. In contrast, the 'black-box' nature of LSTM networks makes it challenging to understand the internal decision-making process.

## 2.4 Proposing Self-Regulating LSTM (SR-LSTM)

SR-LSTM, or Self-Regulating LSTM, enhances the standard LSTM architecture by integrating a self-regulation mechanism. This innovative addition incorporates an extra regulatory layer within the LSTM cell. The primary function of this layer is to dynamically modify the cell's learning behaviours and memory retention capabilities. It does so by continuously adapting to the changing context of the input data sequence. Its adaptive nature enables the SR-LSTM to more effectively manage and process varying data patterns, potentially leading to more accurate and context-sensitive outcomes in tasks involving sequential data processing.

### Proposed Mathematical Formulation for SR-LSTM

1. **Regulatory Factor Calculation**:

$$\mathbf{R_t} = \rho(W_r \cdot h_{t-1} + b_r)$$

Here, $\mathbf{R_t}$ is the regulatory factor for time step $t$, $W_r$ is the weight matrix, $b_r$ is the bias term, and $\rho$ is a non-linear activation function (e.g., Rectified Linear Units (ReLU)).

2. **Modified Forget and Input Gates**:

$$f'_t = f_t \odot \mathbf{R_t}$$

$$i'_t = i_t \odot \mathbf{R_t}$$

We can modify the standard forget gate $f_t$ and input gate $i_t$ by the regulatory factor $\mathbf{R_t}$ to adjust the information flow dynamically.

3. **Cell State and Output Gate**: The cell state update and output gate equations remain the same as in standard LSTM, but they operate on the modified gates $f'_t$ and $i'_t$.

### Rationale for SR-LSTM

- **Dynamic Adaptation**: The self-regulation mechanism allows the LSTM to dynamically adjust its learning behavior based on the context, which could be particularly useful in scenarios with changing data distributions or non-stationary time series.

- **Enhanced Memory Management**: By regulating the forget and input gates, the SR-LSTM can potentially manage its long-term memory more effectively, deciding when to retain or forget information based on the evolving importance of different data aspects.
- **Application in Complex Environments**: This model could be beneficial in complex environments like financial markets or ecological systems, where the relevance and impact of information can change rapidly.
- **Potential for Improved Generalisation**: The self-regulating feature might help in preventing overfitting by ensuring that the model does not rely too heavily on specific features of the training data, thereby potentially improving generalisation to new data.

**Note:** The SR-LSTM concept is speculative and would require extensive research and experimentation to validate its effectiveness and practical utility.

## 3 Methodology

### 3.1 Data Acquisition

The methodology for this study on utilising LSTM networks for trend analysis and prediction of road traffic accidents in Great Britain begins with a meticulous data collection process. The datasets encompass three primary categories: **collisions**, **casualties**, and **vehicles**, each offering unique insights into the multifaceted nature of road traffic accidents.

**Python Code for Data Collection**

The Python code defines a class `ODSConverter` to handle the downloading of an OpenDocument Spreadsheet (ODS) file and converts selected sheets from this file into CSV format. This process is paramount for data collection, mainly when dealing with large datasets like road traffic accidents, casualties, and vehicle information.

```python
import os
import requests
import pandas as pd

class ODSConverter:
    """
    A class to handle downloading an ODS file
    and converting specified sheets to CSV files.

    Attributes:
    data_folder (str): The folder where the ODS
    file will be downloaded and CSV files will be
    saved.
    """

    def __init__(self, data_folder='data'):
        """
        Initialise the ODSConverter with a
    specified data folder.

        Args:
        data_folder (str): The folder where the
    ODS file will be downloaded and CSV files will
     be saved.
        """
        self.data_folder = data_folder

    def download_file(self, url, filename):
        """
        Download a file from a URL into the
    instance's data folder.

        Args:
        url (str): URL of the file to be
    downloaded.
        filename (str): Name of the file to save
     the download as.
        """
        response = requests.get(url)
        response.raise_for_status()  # Raises
    HTTPError for unsuccessful status codes

        file_path = os.path.join(self.
    data_folder, filename)

        if not os.path.exists(self.data_folder):
            os.makedirs(self.data_folder)

        with open(file_path, 'wb') as file:
            file.write(response.content)

        return file_path

    def export_sheets_to_csv(self, ods_file_path
    , sheets, csv_files):
        """
        Export specified sheets from an ODS file
     to CSV files.

        Args:
        ods_file_path (str): Path to the ODS
    file.
        sheets (list of str): Names of the
    sheets to be exported.
        csv_files (list of str): Names of the
    output CSV files.
        """
        for sheet_name, csv_file in zip(sheets,
    csv_files):
            dataframe = pd.read_excel(
    ods_file_path, engine='odf', sheet_name=
    sheet_name)
            csv_path = os.path.join(self.
    data_folder, csv_file)
```

```
55            dataframe.to_csv(csv_path, index=
    False)
56
57 # Usage
58 converter = ODSConverter()
59
60 # URL of the ODS file
61 ods_url = 'https://assets.publishing.service.gov
    .uk/media/65143a8888281e0014b4e95d/ras0101.ods
    '
62 ods_filename = 'ras0101.ods'
63
64 # Download the ODS file
65 ods_file_path = converter.download_file(ods_url,
     ods_filename)
66
67 # Specify sheets and corresponding CSV files
68 sheets_to_export = ['Collisions', 'Casualties',
    'Vehicles']
69 csv_output_files = ['collisions.csv', '
    casualties.csv', 'vehicles.csv']
70
71 # Export the sheets to CSV
72 converter.export_sheets_to_csv(ods_file_path,
    sheets_to_export, csv_output_files)
```

**Listing 1.** Python code for data collection and processing

1. **Import Necessary Libraries**: The code imports `os` for operating system-dependent functionality, `requests` for HTTP requests to download files, and `pandas` for data manipulation and analysis.
2. **Define the ODSConverter Class**: It defines an attribute `data_folder`, which specifies the directory where the downloaded ODS file and the converted CSV files will be stored. The `__init__` method Initialises the class with the `data_folder` attribute.
3. **Download File Method**:

   ```
   def download_file(self, url, filename):
       ...
   ```

   This method downloads a file from a given URL. It uses the `requests.get` method to fetch the file and write the content to a file in the specified `data_folder`. If the folder does not exist, it creates one.
4. **Export Sheets to CSV Method**:

   ```
   def export_sheets_to_csv(self,...):
       ...
   ```

   This method actively converts specified sheets from the downloaded ODS file into CSV files. It iterates over the provided sheet names and corresponding CSV file names, reads each sheet into a pandas DataFrame using `pd.read_excel`, and subsequently saves each DataFrame as a CSV file.

5. **Usage of ODSConverter**: An instance of `ODSConverter` is created. The URL of the ODS file and the filename to save it as are specified. The `download_file` method is called to download the ODS file. The sheets to be exported and their corresponding CSV filenames are defined. The `export_sheets_to_csv` method is called to convert the specified sheets to CSV files.

This code is particularly useful for automating the process of data collection and conversion, making it efficient to handle large datasets required for analysis in studies like traffic accident trend analysis. By converting the data into CSV format, it becomes easier to manipulate and analyse using a wide range of data analysis tools and libraries.

### 3.2 Understanding the Datasets

Understanding datasets prior to model training is imperative. It guarantees data integrity, preventing flawed, biased predictions from dataset errors. Recognising key features streamlines model creation, as irrelevant features diminish performance. Such comprehension is necessary in choosing the optimal model for specific data types and tasks, thus boosting effectiveness. It also fosters efficient training, saving computational resources and time through adept data preprocessing. Finally, data insight is fundamental in adhering to legal and ethical norms, particularly with sensitive data, securing the model's responsible deployment.

**Collisions Dataset**: This dataset offers a detailed account of each reported road traffic collision, encompassing a range of factors that provide a comprehensive understanding of each incident. The dataset includes the following key columns:

- **Year**: This column records the year the collision occurred. It is crucial to analyse trends and understand how collision patterns have evolved.
- **Fatal**: This field denotes the presence of fatalities in a collision, indicated solely by numeric values (e.g., 1474, 1391). It is an indispensable metric for distinguishing between accidents with and without fatalities. Analysing these figures is required to gauge accidents' gravity and craft precise safety interventions.
- **FSC Unadjusted**: Standing for **Fatal or Serious Collisions Unadjusted**, this column represents the count of collisions that resulted in fatal or severe injuries, without adjustments for under-reporting or other factors. This raw data is essential for initial assessments of road safety conditions.
- **FSC Adjusted**: Contrasting with the previous column, **Fatal or Serious Collisions Adjusted** accounts for under-reporting and other known discrepancies in the collision data. This adjusted figure provides a more accurate representation of serious incidents, offering a reliable basis for policy-making and safety interventions.
- **All Collisions**: This column encompasses the total number of collisions reported, regardless of severity. It includes all incidents, from minor to fatal, providing a comprehensive overview of the overall road safety situation.

Each record in the dataset corresponds to a unique collision event, tagged with a unique identifier for cross-referencing with the casualties and vehicles datasets. The data from the Department for Transport (DfT) covers 1926 to 2022. These meticulous categorisation and recording details ensure a robust foundation for analysing collision trends and formulating effective road safety strategies.

**Casualties Dataset**: This dataset delves into the human impact of road traffic accidents, providing detailed information about the casualties involved. It includes several columns that help in understanding the extent and nature of injuries sustained in these incidents:
- **Year**: This column records the year of the accident, which is necessary for temporal analysis and identifying trends in casualty rates over time.
- **Pedestrians Killed**: This field indicates the number of pedestrians who died in road traffic accidents. It is required to assess pedestrian safety and the effectiveness of measures to protect this vulnerable group.
- **Pedal Cyclists Killed**: This column shows the number of pedal cyclists killed in collisions. It highlights cyclists' risks and is vital for developing targeted safety initiatives for this group.
- **Motorcyclists Killed**: This field represents the number of motorcyclists fatally injured in accidents. The data is critical to understanding motorcyclists' risks and crafting policies to enhance their safety.
- **Car Occupants Killed**: This column details the number of fatalities among car occupants. It provides insights into the safety of cars and the effectiveness of in-vehicle safety features.
- **Other Road Users Killed**: This category includes fatalities among other types of road users not covered in the above categories, such as truck drivers, bus passengers, etc. It ensures a comprehensive understanding of all road user fatalities.
- **All Road Users Killed**: This field sums up the total fatalities across all categories of road users. It provides an overall picture of the lethality of road traffic accidents in a given year.
- **All Road Users All Severities**: Unlike the previous column, this one accounts for all casualties, regardless of the severity of their injuries. It includes minor, serious, and fatal injuries, offering a complete overview of the impact of road traffic accidents on people.

Each record in the casualties dataset is meticulously recorded and aligned with the collision dataset regarding time frame and unique identifiers, enabling a comprehensive analysis of the correlation between collision events and casualty outcomes. This dataset, obtained from the Department for Transport, is instrumental in understanding the human cost of road traffic accidents and identifying patterns or trends in casualty rates over time.

**Vehicles Dataset**: This dataset complements the collisions and casualties datasets by providing detailed information on the vehicles involved in each accident. It encompasses a variety of vehicle types, offering insights into the role of different vehicles in road traffic accidents. The columns in this dataset include:
- **Year**: This column records the accident's year, indispensable for trend analysis and understanding how different types of vehicles contribute to road safety issues.
- **Pedal Cycles**: This field indicates the number of pedal cycles (bicycles) involved in road accidents. It is crucial to assess the safety of cyclists and the interaction between bicycles and other vehicles on the road.
- **Motorcycles**: This column shows the number of motorcycles involved in collisions. The data is necessary for understanding motorcyclists' risks and formulating safety measures specific to this category of road users.
- **Cars**: This field details the number of cars involved in accidents. It provides insights into the prevalence of car involvement in road accidents and is vital for evaluating car safety standards and driver behaviour.
- **Buses or Coaches**: This column represents the number of buses or coaches involved in road traffic accidents. It helps in assessing the safety of public transport vehicles and their passengers.
- **Light Goods Vehicles**: This includes light commercial vehicles such as vans and small trucks. The data helps us understand the role of commercial transport in road safety.
- **Heavy Goods Vehicles**: This field covers heavy goods vehicles like large trucks and lorries. It is crucial to analyse the impact of heavy commercial vehicles on road safety and develop strategies to mitigate their risk.
- **Other Vehicles**: This category encompasses other types of vehicles not covered in the above categories, such as tractors, quad bikes, etc. It ensures a comprehensive understanding of all vehicle types involved in road accidents.
- **Unknown Vehicles**: This column accounts for vehicles whose type could not be determined or the system

did not record. Including this data helps maintain the integrity of the dataset and acknowledges the presence of data gaps.
- **All Vehicles**: This field sums up the total number of vehicles across all categories involved in road accidents. It provides an overall picture of vehicle involvement in road traffic accidents for any given year.

Each vehicle dataset record is meticulously recorded and aligned with the collisions and casualties datasets, offering a holistic view of each accident event, encompassing environmental, human, and vehicular factors. This dataset, sourced from the Department for Transport, is instrumental in assessing the role of different vehicle types in road traffic accidents and formulating comprehensive road safety strategies.

### 3.3 Data Preprocessing

Data preprocessing is necessary to ensure the quality and reliability of the data used for analysis. This study focuses on preprocessing datasets related to casualties, collisions, and vehicles. The preprocessing involves cleansing and structuring the data, which includes removing unnecessary rows, adding appropriate headers, and cleansing specific fields to ensure accuracy and consistency. The following Python code illustrates the implementation of the data preprocessing steps using a custom class, `DataPreprocessor`.

```python
import pandas as pd

class DataPreprocessor:
    """
    A class to handle preprocessing of data files.

    This class provides methods to preprocess
    casualties, collisions, and vehicles data
    files.
    It removes specified rows, adds headers,
    cleans specific fields, and saves the
    processed files.
    """

    def __init__(self, data_folder='data'):
        """
        Initialise the DataPreprocessor with a
    specified data folder.

        Args:
        data_folder (str): The folder where the
    data files are located and processed files
    will be saved.
        """
        self.data_folder = data_folder

    def preprocess_file(self, file_name,
    skip_rows, new_headers, output_file):
        """
        Preprocess a data file: remove specific
    rows, add new headers, clean fields, and save
    the processed file.

        Args:
        file_name (str): Name of the file to be
    preprocessed.
        skip_rows (int): Number of initial rows
    to skip.
        new_headers (list of str): New headers
    to be added to the file.
        output_file (str): Name of the output
    file for the processed data.
        """
        file_path = f'{self.data_folder}/{
    file_name}'
        df = pd.read_csv(file_path, skiprows=
    skip_rows, header=None)
        df = df.iloc[:, :len(new_headers)]  #
    Select only the columns we need
        df.columns = new_headers
        df.replace(to_replace=r'\[.*\]', value='
    ', regex=True, inplace=True)
        output_path = f'{self.data_folder}/{
    output_file}'
        df.to_csv(output_path, index=False)

# Usage
preprocessor = DataPreprocessor()

# Define new headers
casualties_headers = ["year", "
    pedestrians_killed", "pedal_cyclists_killed",
    "motocyclists_killed",
                      "car_occupants_killed", "
    other_road_users_killed", "
    all_road_users_killed",
                      "
    all_road_users_all_severities"]

# Preprocess casualties.csv
preprocessor.preprocess_file('casualties.csv',
    7, casualties_headers, 'casualties_cleansed.
    csv')

# Preprocess collisions.csv
collisions_headers = ["year", "fatal", "
    fsc_unadjusted", "fsc_adjusted", "
    all_collisions"]
preprocessor.preprocess_file('collisions.csv',
    6, collisions_headers, 'collisions_cleansed.
    csv')

# Preprocess vehicles.csv
vehicles_headers = ["year", "pedal_cycles", "
    motorcycles", "cars", "buses_or_coaches",
                    "light_goods_vehicles", "
    heavy_goods_vehicles", "other_vehicles",
```

```
56                         "unknown_vehicles", "
           all_vehicles"]
57    preprocessor.preprocess_file('vehicles.csv', 4,
          vehicles_headers, 'vehicles_cleansed.csv')
```

**Listing 2.** Python code for data preprocessing

### 3.4 Model Training Process

The training process for the LSTM model in this study involves several steps, including handling **NaN** values and selecting model parameters. A pivotal decision in this process is the scaling method for the casualties, collisions, and vehicle datasets, which hinges on the nature of the data, particularly concerning the presence of outliers.

**Scaling Method Selection**: The decision between using MinMaxScaler and RobustScaler from sklearn is compulsory. MinMaxScaler, effective when data lacks significant outliers, scales each feature to a specified range, typically [0, 1]. This scaler is ideal for transforming features to a standard scale but is sensitive to outliers. Therefore, if the casualties, collisions, and vehicle datasets contain outliers, MinMaxScaler may not be the best choice. On the other hand, RobustScaler utilises statistics robust to outliers—specifically, the median and the interquartile range (IQR) [2] [14]. It scales features using these statistics, making it less sensitive to outliers. Given the nature of road traffic data, where outliers can be expected, especially in accident data with extreme values like unusually high casualties or rare vehicle collisions, RobustScaler is preferable.

**Data Distribution Considerations**: The data distribution also plays a crucial role in the choice of scaler. RobustScaler is a better option for datasets not normally distributed due to its reliance on the median and IQR, which are less sensitive to non-normal distributions.

**Final Decision for Scaling**: For the casualties, collisions, and vehicle datasets, which are likely to contain outliers due to the nature of road traffic incidents, such as rare but severe accidents, the RobustScaler is generally more appropriate. It scales features without being unduly influenced by extreme values. However, thoroughly analysing the specific characteristics of the datasets in question should guide the final decision on the scaling method.

In the subsequent steps of the methodology, our focus will shift to handling NaN values and determining the optimal parameters for the LSTM model, ensuring the model is well-tuned to the nuances of the traffic accident data.

---

[2]"When a data set has outliers, variability is often summarized by a statistic called the interquartile range, which is the difference between the first and third quartiles."

### 3.5 Installing Python Libraries

Installing the necessary libraries ensures that the Python code used in this study executes appropriately. The following commands are executed in a Python 3.10.5 environment to install the latest versions of libraries such as **tqdm**, **numpy**, **pandas**, **matplotlib**, **tensorflow**, **scikit-learn**, and **transformers**. These libraries provide a range of functionalities, from data manipulation and visualisation to machine learning and deep learning capabilities.

```
1   !pip install --upgrade pip
2   !pip install -q -U tqdm
3   !pip install -q -U numpy
4   !pip install -q -U pandas
5   !pip install -q -U matplotlib
6   !pip install -q -U tensorflow
7   !pip install -q -U scikit-learn
8   !pip install -q -U transformers
```

**Listing 3.** Python code for installing libraries

### 3.6 Importing Python Libraries

This study's initial step involves importing a suite of Python libraries. These libraries provide the tools and functions for data analysis, preprocessing, model development, and evaluation.

```
1   import numpy as np
2   import pandas as pd
3   import matplotlib.pyplot as plt
4   from sklearn.model_selection import
        train_test_split
5   from sklearn.preprocessing import RobustScaler
6   from tensorflow.keras.models import Sequential
7   from tensorflow.keras.layers import LSTM, Dense,
        Dropout
8   from tensorflow.keras.callbacks import
        EarlyStopping, ModelCheckpoint
9   from sklearn.metrics import mean_squared_error,
        mean_absolute_error
```

**Listing 4.** Python code for importing libraries

### 3.7 Converting Strings to Floats

Data processing often requires converting string values to numerical formats for analysis. We wrote a Python function for converting string values to floating-point numbers. The **convert_to_float** function robustly handles the conversion, ensuring that any invalid string values, which cannot directly convert to floats, default to zero. Handling invalid values by defaulting them to zero maintains data integrity and avoids errors during numerical computations.

```
1   def convert_to_float(x):
2       try:
```

```
3            return float(str(x).replace(',', '').strip
         ())
4    except ValueError:
5        return 0
```
**Listing 5.** Python function for converting strings to floats

### 3.8 Data Loading and Cleaning

The datasets include information on collisions, casualties, and vehicles, each cleansed and prepared for analysis. The cleansing steps involve applying a conversion function to ensure data consistency, handling **NaNs** and **infinities**, merging datasets based on the **year** column, and scaling the features using **RobustScaler**. These steps are necessary to prepare the data for subsequent analysis and ensure the integrity and reliability of our models.

```
1  # Load the datasets
2  collisions = pd.read_csv('collisions_cleansed.csv'
       )
3  casualties = pd.read_csv('casualties_cleansed.csv'
       )
4  vehicles = pd.read_csv('vehicles_cleansed.csv')
5
6  # Apply the conversion function to each DataFrame
       using DataFrame.apply
7  collisions = collisions.apply(lambda col: col.
       apply(convert_to_float))
8  casualties = casualties.apply(lambda col: col.
       apply(convert_to_float))
9  vehicles = vehicles.apply(lambda col: col.apply(
       convert_to_float))
10
11 # Handle NaNs and infinities in the datasets
       before scaling
12 for df in [collisions, casualties, vehicles]:
13     df.replace([np.inf, -np.inf], np.nan, inplace=
       True)  # Replace infinities with NaN
14     df.fillna(df.median(), inplace=True)  #
       Replace NaNs with median
15
16 # Define the relevant columns that need conversion
17 relevant_columns = ['all_collisions', '
       all_road_users_killed', '
       all_road_users_all_severities',
18                     'pedal_cycles', 'motorcycles',
        'cars', 'buses_or_coaches',
19                     'light_goods_vehicles', '
       heavy_goods_vehicles', 'other_vehicles',
20                     'unknown_vehicles', '
       all_vehicles']
21
22 # Merge the datasets on 'year' column
23 merged_data = collisions.merge(casualties, on='
       year', how='outer').merge(vehicles, on='year',
        how='outer')
24
25 # Select features and target variable
26 features = merged_data[relevant_columns]
27 target = merged_data['all_road_users_killed']
28
29 # Use RobustScaler for features
30 robust_scaler = RobustScaler()
31 scaled_features = robust_scaler.fit_transform(
       features)
32
33 # Replace any NaN values produced by scaling with
       0
34 scaled_features = np.where(np.isnan(
       scaled_features), 0, scaled_features)
35
36 # Recheck for NaN or Infinite values after scaling
37 if np.any(np.isnan(scaled_features)) or np.any(np.
       isinf(scaled_features)):
38     raise ValueError("Data contains NaN or
       Infinite values after scaling.")
```
**Listing 6.** Python code for data loading and cleaning

### 3.9 Building The Model

Following the data loading and cleaning steps, the next phase in our methodology involves splitting the data into training and testing sets, building the LSTM model, compiling it, training it with validation data, and finally evaluating the model and making predictions. This exhaustive approach ensures the development of a robust model capable of accurate trend analysis and prediction of road traffic accidents.

**Splitting Data into Training and Testing Sets**
The first step post-cleaning is to split the datasets into training and testing sets. This division is compulsory for evaluating the model's performance on unseen data as it ensures its generalisability. We typically allocate **70-80%** of the data for training and the remainder for testing. The **train_test_split** function from **sklearn.model_selection** facilitates this process. Maintaining a balance between the training and testing sets is integral to avoid overfitting or underfitting the model.

**Building the LSTM Model**
The LSTM model's architecture is pivotal in capturing the temporal dependencies in the traffic data. We use the **Sequential** model from **tensorflow.keras.models** to create a stack of layers. The model comprises multiple LSTM layers, interspersed with **Dropout** layers to prevent overfitting. Each LSTM layer, created using the **LSTM** class from **tensorflow.keras.layers**, has parameters such as the number of units and activation function, which are crucial for learning the data's temporal features. The concluding layer in our model is a singular-unit **Dense** layer designed explicitly for regression output. This configuration is essential for predicting continuous variables, such as the incidence rate of traffic accidents.

**Compiling the Model**
Once the model architecture is defined, we compile the model using an appropriate optimiser and loss function. The choice of optimiser, such as Adam or RMSprop, affects the speed and quality of the learning process—in our case, we utilised Adam. For regression tasks like ours, the mean squared error (MSE) is a common choice for the loss function. The **compile** method of the Sequential model is used for this purpose.

**Training the Model with Validation Data**
Training the model involves feeding the training data into the model and adjusting the model weights to minimise the loss function. We use the **fit** method of the Sequential model, indicating the quantity of epochs and the size of each batch. Additionally, we use a portion of the training data as validation data. This approach allows us to monitor the model's performance on unseen data during training, providing an early indication of overfitting. The **EarlyStopping** callback from **tensorflow.keras.callbacks** is employed to stop training when the validation loss ceases to decrease, further preventing overfitting.

**Model Evaluation and Making Predictions**
After training, we evaluate the model's performance on the testing set. This evaluation is crucial to understand how well the model generalises to new data. We employed metrics like the MAE[3] and RMSE[4] to gain insights into the accuracy of the model's predictions.

The **evaluate** method of the Sequential model, along with the **mean_squared_error** and **mean_absolute_error** functions from **sklearn.metrics**, are used for this purpose. Finally, we deploy the trained model to generate predictions. We apply the model's **predict** method to forecast future values using the testing set. We then compare these predictions with the actual values to assess the model's effectiveness.

In summary, this comprehensive approach—encompassing data splitting, model building, compilation, training, and evaluation—ensures the development of a robust LSTM model. This model can effectively analyse and predict trends in road traffic accidents, thereby providing valuable insights for road safety measures and strategies.

## 4 Results
### 4.1 Presentation of the LSTM Model's Performance

We quantitatively evaluated the performance of the Long Short-Term Memory (LSTM) model in this study using two primary metrics: Root Mean Squared Error (RMSE) and Mean Absolute Error (MAE). These metrics play an integral role in assessing the accuracy of the model's predictions compared to the actual values.

**Root Mean Squared Error (RMSE)**: RMSE is a standard quantitative data analysis metric for quantifying a model's prediction error. We define it as the square root of the mean of the squared differences between predicted and actual values. [16] The mathematical representation of RMSE is as follows:

$$\text{RMSE} = \sqrt{\frac{1}{n} \sum_{i=1}^{n}(y_i - \hat{y}_i)^2}$$

where $y_i$ represents the actual values, $\hat{y}_i$ denotes the predicted values, and $n$ is the number of observations. In our study, the RMSE for the LSTM model is 4206.463547126862.

**Mean Absolute Error (MAE)**: MAE measures the average magnitude of the errors in a set of predictions, without considering their direction. The MAE is given by:

$$\text{MAE} = \frac{1}{n} \sum_{i=1}^{n} |y_i - \hat{y}_i|$$

For our LSTM model, the MAE computes as 3734.0478454589843.

### 4.2 Interpretation of the LSTM Model Results

The obtained RMSE and MAE values offer significant insights into the LSTM model's performance. An RMSE of 4206.464 indicates that the model's predictions are, on average, within this range of the actual values. The MAE of 3734.048 suggests that the average prediction error, disregarding the error direction, is within this range.

RMSE and MAE are crucial in evaluating the model's performance, but they also have limitations. RMSE is more sensitive to outliers than MAE, as it squares the errors before averaging, thus giving a relatively high weight to significant errors [17]. This sensitivity can be advantageous or disadvantageous depending on the specific application and the nature of the data.

In contrast, MAE provides a more balanced view of the model's performance across all predictions, treating all errors equally. It proves beneficial when outliers do not disproportionately impact the data or when seeking a more conservative assessment of the model's performance.

In conclusion, the LSTM model demonstrates a reasonable level of accuracy in predicting road traffic accidents, as indicated by the RMSE and MAE values. Further refinement of the model may be necessary to improve its predictive accuracy, particularly in reducing the average error magnitude as indicated by these metrics.

---

[3]MAE (Mean Absolute Error) measures errors between paired observations expressing the same phenomenon.
[4]RMSE (Root Mean Squared Error) measures the differences between values predicted by a model and the values observed. [15]

## 4.3 Trend Analysis Over Time

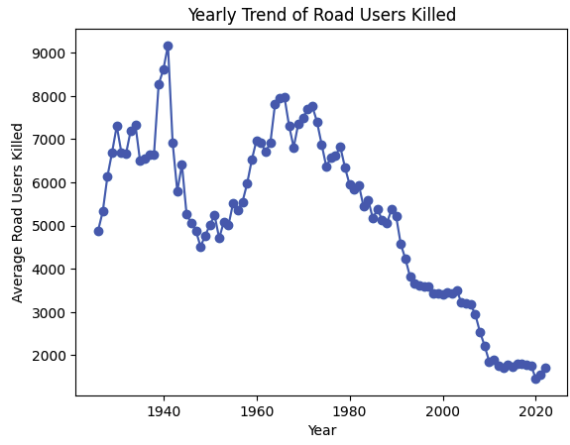

**Figure 1.** Yearly Trend of Road Users Killed

Our analysis thoroughly examines the trend in annual road fatalities from before 1929 to 2022, using a dataset aggregated by year. The data shows a significant decrease in road user deaths over time, represented graphically with a line graph marked in blue.

During the 1940s to late 1960s, fluctuations and peaks in fatalities were observed, with a notable peak suggesting an outlier year. Post-1960s, fatalities plateaued, then steadily declined from the 1970s onwards, correlating with global road safety initiatives like seat belt laws and improved vehicle standards. This decline stabilised in the 1980s and 1990s.

In the 21st century, the trend continues downward, reaching the lowest average fatalities attributed to advancements in automotive technology and road safety campaigns. The 2020s show a significant dip due to COVID-19 pandemic restrictions reducing vehicular traffic.

The trend analysis indicates that various factors, including technological, legislative, and societal changes, influence the decrease in fatalities. It employs a mean annual fatality measure, smoothing out short-term fluctuations to highlight long-term trends.

The analysis shows a substantial reduction in road fatalities over the past eight decades, influenced by technology, policy, and behavioural changes. We concluded and linked the 2020 decrease to COVID-19's impacts on transportation. This study accentuates the importance of continued investment in safety initiatives and the need to monitor future socio-economic impacts on road safety.

## 4.4 Correlation of Vehicle Types

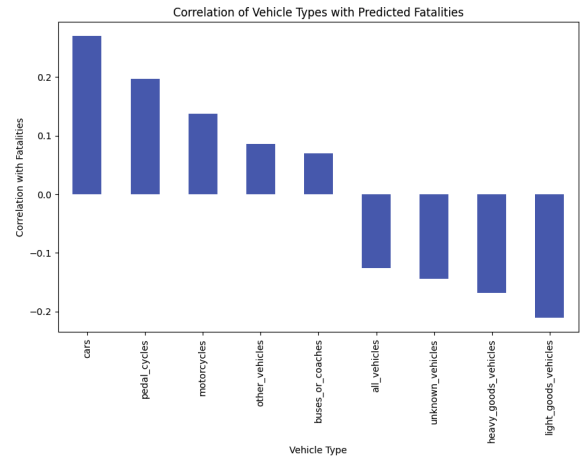

**Figure 2.** Correlation of Vehicle Types with Predicted Fatalities

The data-driven approach to road safety regulation necessitates focusing on vehicles exhibiting the highest correlation with fatalities. For cars, which are the predominant contributors to road fatalities, the integration of advanced safety features is paramount—evidence supports the adoption of technologies such as electronic stability control (ESC) and autonomous emergency braking (AEB) to reduce collision rates significantly (Lie et al., 2008) [18]. Moreover, regular safety campaigns actively reinforce the importance of seatbelt use and have proven effective in enhancing compliance and reducing injuries (Williams & Wells, 2004) [19].

Pedal cycles and motorcycles, characterised by vulnerability, require interventions tailored to their exposure. Helmet use, for instance, has been consistently associated with reductions in head injury risk (Thompson et al., 1996) [20]. Investment in cycling infrastructure can decrease collision rates, providing a safer environment for cyclists (Reynolds et al., 2009) [21].

For larger vehicles such as buses and coaches, the emphasis must be on professional training for drivers and vehicle maintenance. Research has indicated that driver error significantly contributes to bus-related incidents, and comprehensive driver safety programs can mitigate this risk. (Transport Research Laboratory, 1996) [22].

The negative correlation observed for specific vehicle categories suggests effective existing regulations or under-reporting. For heavy and light goods vehicles, stringent licencing and operational standards correlate with lower accident rates. (Aarts & van Schagen, 2006) [23].

Regulatory bodies must continue adapting road safety measures to align with statistical trends, effectively reducing road traffic fatalities. Additionally, research supports these strategies, and continuous evaluation in response to new data and technological advances is indispensable.

## 5 Conclusion

### 5.1 Summary of the Findings from the Result

This study's exploration into applying Long Short-Term Memory (LSTM) networks for predicting road traffic accidents in Great Britain yielded significant findings. We evaluated the LSTM model's performance using Root Mean Squared Error (RMSE) and Mean Absolute Error (MAE), demonstrating a reasonable accuracy level. With an RMSE of 4206.464 and an MAE of 3734.048, the model effectively captured the variability and trends within the traffic accident data. Whilst highlighting the model's predictive capabilities, these metrics underscore the complexity and challenges of forecasting road traffic accidents.

The trend analysis over time revealed a notable decrease in road fatalities, reflecting the impact of various safety measures and technological advancements in vehicles. This decline, particularly evident from the 1970s onwards, aligns with global road safety initiatives and improvements in vehicle safety standards. The analysis also pointed out the significant reduction in fatalities during the 2020s, likely due to reduced vehicular traffic amid COVID-19 pandemic restrictions.

The correlation study of vehicle types with predicted fatalities emphasised the importance of focusing on specific vehicle categories that contribute most to road fatalities. Cars, being the predominant contributors, require the integration of advanced safety features, whilst vulnerable vehicle types like pedal cycles and motorcycles necessitate tailored safety interventions.

### 5.2 Implications for Future Road Safety Measures

The findings of this study have profound implications for future road safety measures. The demonstrated effectiveness of LSTM in predicting road traffic accidents suggests that such models can be integral in formulating and evaluating road safety strategies. Policymakers and road safety authorities can leverage these insights to prioritise areas requiring immediate attention and allocate resources more effectively.

The trend analysis emphasises the necessity for ongoing investment in road safety initiatives, especially considering the evolution of vehicle technologies and shifting traffic patterns. This study strongly advocates for a data-driven approach to road safety, where decision-making relies on insights from predictive analytics and trend analysis.

### 5.3 Proposition of Self-Regulating LSTM (SR-LSTM)

In light of the study's findings, we propose developing a Self-Regulating LSTM (SR-LSTM) model. This advanced model aims to enhance the predictive accuracy of standard LSTM by incorporating a self-regulation mechanism. The SR-LSTM would dynamically adjust its learning behaviour based on the context, offering a more nuanced understanding of traffic patterns and potentially improving prediction accuracy. This proposition opens new avenues for research and development in traffic accident prediction and road safety analysis.

### 5.4 Limitations of the Current Research

Whilst the study provides valuable insights, it has limitations. The primary constraint is the model's sensitivity to outliers, as the RMSE metric indicates. Although beneficial in capturing extreme values, this sensitivity might sometimes overemphasise rare but severe accidents. Additionally, the study relies on historical data, which may need to fully encapsulate future changes in traffic behaviour, vehicle technology, and road safety policies.

Another area for improvement lies in the data's scope, focusing primarily on Great Britain. Extending this research to include data from other regions could provide a more comprehensive understanding of global traffic patterns and accident trends.

In conclusion, the study represents a significant step in applying LSTM networks for road traffic accident prediction. The findings demonstrate the model's potential in predictive analytics and highlight areas for future research and development. The proposed SR-LSTM model and the insights gained from this study can significantly contribute to enhancing road safety measures and strategies. Continuous advancements in this field and a data-driven approach are necessary to reduce road traffic accidents and improve public safety.

## 6 Appendices

### 6.1 Appendix A: Data Availability

The datasets generated and/or analysed during the study are available on gov.uk.

### 6.2 Appendix B: Jupyter Notebook

The Jupyter Notebook for training the model and conducting the analysis is available here.

## 7 Ethical Standard

The research meets all ethical guidelines, including adherence to the legal requirements of the United Kingdom.

## 8 Acknowledgements

Our profound thanks go to all contributors to this research project's success. We particularly acknowledge colleagues at the Driver and Vehicle Standards Agency (DVSA), Driver and Vehicle Licensing Agency (DVLA), and Department for Transport (DfT) for their guidance, astute feedback, and steadfast support during our analysis. Their dedication to road safety in Great Britain is remarkable and deserves our heartfelt appreciation.

The success of this work hinges directly on the efforts and support of the individuals named and numerous others engaged in this project.

## References

[1] Department for Transport. Reported road casualties in great britain: 2019 annual report, 2020.
[2] Sepp Hochreiter and Jürgen Schmidhuber. Long short-term memory. *Neural Computation*, 9(8):1735–1780, 1997.
[3] Shaw-Pin Miaou and H Lum. Modeling vehicle accidents and highway geometric design relationships. *Accident Analysis & Prevention*, 25(6):689–709, 1993.
[4] James D. Hamilton. *Time Series Analysis*. Princeton University Press, 1994.
[5] Sepp Hochreiter and Jürgen Schmidhuber. Long short-term memory. *Neural Computation*, 9(8):1735–1780, 1997.
[6] Felix A Gers, Jürgen Schmidhuber, and Fred Cummins. Learning to forget: Continual prediction with lstm. *Neural Computation*, 12(10):2451–2471, 2000.
[7] Sima Siami-Namini, Neda Tavakoli, and Akbar Siami Namin. The performance of lstm and bilstm in forecasting time series. In *2018 IEEE International Conference on Big Data (Big Data)*, pages 3285–3292. IEEE, 2018.
[8] Xingjian Shi, Zhourong Chen, Hao Wang, Dit-Yan Yeung, Wai-kin Wong, and Wang-chun Woo. Convolutional lstm network: A machine learning approach for precipitation nowcasting. *Advances in Neural Information Processing Systems*, 30, 2017.
[9] Carnegie Mellon University Machine Learning Department. Explaining a black box using deep variational information bottleneck approach, 2019. Accessed: 30/11/2023.
[10] Bing Yu, Haoteng Yin, and Zhanxing Zhu. Spatio-temporal graph convolutional networks: A deep learning framework for traffic forecasting. In *Proceedings of the 27th International Joint Conference on Artificial Intelligence*, pages 3634–3640, 2017.
[11] Junbo Deng and Hongzi Zhao. Lstm-based traffic flow prediction with missing data. *Neurocomputing*, 318:297–305, 2018.
[12] Sepp Hochreiter and Jürgen Schmidhuber. Long short-term memory. *Neural Computation*, 9(8):1735–1780, 1997.
[13] George EP Box and Gwilym M Jenkins. *Time series analysis: Forecasting and control*. Holden-Day, 1976.
[14] Lisa Sullivan and Wayne W. LaMorte. When a data set has outliers, variability is often summarized by a statistic called the interquartile range, which is the difference between the first and third quartiles, 2016.
[15] Masoud Goharimanesh, Ehsan Abbasnejad Jannatabadi, Hossein Moeinkhah, Mohammad Bagher Naghibi Sistani, and Ali Akbar Akbari. An intelligent controller for ionic polymer metal composites using optimized fuzzy reinforcement learning. *Journal of Intelligent and Fuzzy Systems*, 2017.
[16] Lei Wang et al. Artificial intelligence model integrated with bim model for core construction of transportation hub. *International Journal of Recent Innovation in Trends in Computing and Communication*, 11(6S), 2023.
[17] Cort J. Willmott and Kenji Matsuura. Advantages of the mean absolute error (mae) over the root mean square error (rmse) in assessing average model performance. *Climate Research*, 30:79–82, 2005.
[18] Anders Lie, Anders Kullgren, and Claes Tingvall. The effectiveness of electronic stability control (esc) in reducing real life crashes and injuries. *Traffic Injury Prevention*, 9(1):38–43, 2008.
[19] Allan F. Williams and J. K. Wells. The role of enforcement programs in increasing seat belt use. *Journal of Safety Research*, 35(2):175–180, 2004.
[20] Diane C. Thompson, Frederick P. Rivara, and Robert S. Thompson. Effectiveness of bicycle safety helmets in preventing head injuries. *JAMA*, 276(24):1968–1973, 1996.
[21] Conor CO Reynolds, M. Anne Harris, Kay Teschke, Peter A. Cripton, and Meghan Winters. The impact of transportation infrastructure on bicycling injuries and crashes: A review of the literature. *Environmental Health*, 8:47, 2009.
[22] Transport Research Laboratory. Buses involved in road accidents. Report 286, TRL Limited, Crowthorne, 1996.
[23] Leo Aarts and Ingrid van Schagen. Driving speed and the risk of road crashes: A review. *Accident Analysis & Prevention*, 38(2):215–224, 2006.